\documentclass{bmvc2k}

\usepackage{multirow}
\usepackage{makecell}
\usepackage{bbding}
\usepackage{amssymb}
\usepackage{amsmath}

\title{A Hybrid Framework Bridging CNN and ViT based on Theory of Evidence for Diabetic Retinopathy Grading}

\addauthor{Junlai Qiu}{qjl@hainanu.edu.cn}{1, 2}
\addauthor{Yunzhu Chen}{chenyunzhu.cyz@foxmail.com}{3}
\addauthor{Hao Zheng}{howzheng@tencent.com}{4}
\addauthor{Yawen Huang}{yawenhuang@tencent.com}{4}
\addauthor{Yuexiang Li}{yuexiang.li@ieee.org}{$\dagger$, 2}

\addinstitution{
 School of Biomedical Engineering\\
 Hainan University, Sanya, China
}
\addinstitution{
 Medical AI ReSearch (MARS) Group\\
 University Engineering Research Center of Digital Medicine and Healthcare, Guangxi Medical University\\
 Nanning, China
}
\addinstitution{
 School of Information Engineering\\
 Guangxi Polytechnic of Construction\\
 Nanning, China
}
\addinstitution{
 Tencent Jarvis Lab\\
 Tencent, Shenzhen, China
}

\runninghead{Qiu et al.}{Hybrid Framework based on Evidence Theory for DR Grading}

\begin{document}
\renewcommand{\thefootnote}{}
\footnotetext{\hspace{-1.9em}$\dagger$ Corresponding author }
% \thanks{$\dagger$ Corresponding author}
\maketitle

\begin{abstract}
Diabetic retinopathy (DR) is a leading cause of vision loss among middle-aged and elderly people, which significantly impacts their daily lives and mental health. To improve the efficiency of clinical screening and enable the early detection of DR, a variety of automated DR diagnosis systems have been recently established based on convolutional neural network (CNN) or vision Transformer (ViT). However, due to the own shortages of CNN / ViT, the performance of existing methods using single-type backbone has reached a bottleneck. One potential way for the further improvements is integrating different kinds of backbones, which can fully leverage the respective strengths of them (\emph{i.e.,} the local feature extraction capability of CNN and the global feature capturing ability of ViT). To this end, we propose a novel paradigm to effectively fuse the features extracted by different backbones based on the theory of evidence. Specifically, the proposed evidential fusion paradigm transforms the features from different backbones into supporting evidences via a set of deep evidential networks. With the supporting evidences, the aggregated opinion can be accordingly formed, which can be used to adaptively tune the fusion pattern between different backbones and accordingly boost the performance of our hybrid model. We evaluated our method on two publicly available DR grading datasets. The experimental results demonstrate that our hybrid model not only improves the accuracy of DR grading, compared to the state-of-the-art frameworks, but also provides the excellent interpretability for feature fusion and decision-making.
\end{abstract}

%-------------------------------------------------------------------------
\section{Introduction}
\label{sec:intro}
Diabetic retinopathy (DR), a leading cause of vision loss among the middle-aged and elderly people in many countries, not only severely affects the quality of patient lives but also has a significant impact on their mental health \cite{Diabetic_Wong_2016}. 
% DR patients exhibit significant differences in lesion characteristics and clinical manifestations at different stages. In the early stages, the mild symptoms are often overlooked by patients. 
The early detection of DR is pivotal in clinical practice, since
% However, 
the high-grade DR often causes pathological and irreversible changes, such as retinal vascular rupture, obstruction and abnormal proliferation, which eventually cause vision impairment and blindness \cite{MPLNet_Yining_2024,Deep_Nikos_2021}. 
Hence, recent studies \cite{CANet_Li_2020,Deep_Nikos_2021} have proposed a series of deep learning models to perform accurate-and-automated DR grading, which alleviates the workload of ophthalmologists and improves their clinical diagnostic efficiency.
% DR grading provides clinicians with a means to assess the severity of the condition and guide them in developing appropriate treatment plans, which is crucial for preventing further vision loss and blindness. 
% Given the large number of diabetic patients and the high incidence of DR, ophthalmologists are under tremendous work pressure. Therefore, the use of computer-assisted diagnostic technology is essential for alleviating the workload of doctors and improving diagnostic efficiency and accuracy.
% In recent years, the proposal of a series of deep learning models has shown great potential in the field of automated DR grading \cite{Deep_Nikos_2021}. Convolutional Neural Networks (CNNs) are favored for their simple structure, minimal parameter count, ease of training, and robust capability to capture local image features \cite{VanillaNet_Chen_2023}. 
For examples, Li \emph{et al.} \cite{CANet_Li_2020} proposed a cross-disease attention network to jointly grade DR and diabetic macular edema by exploring the internal relationships between diseases using only image-level supervision. Nevertheless, most of existing studies are established upon pure convolutional neural network (CNN) \cite{VanillaNet_Chen_2023} or vision Transformer (ViT) \cite{PVT_Wang_2022,Improving_Yuexiang_2024}. Since either CNN or ViT has its own shortages, \emph{i.e.,} CNN lacks of the capacity to capture long-range dependencies and the ability of local feature extraction of ViT is unsatisfactory, 
% Vision Transformer (ViT), leveraging its self-attention mechanism, can effectively capture global dependencies and optimize the modeling of global context, thereby demonstrating superior performance in visual tasks such as image classification, localization, and segmentation \cite{PVT_Wang_2022}. Li \emph{et al.} proposed the Bootstrap Own Latent of Transformer and pre-trained it on a large-scale fundus image dataset, thereby achieving high-precision performance in DR grading \cite{Improving_Yuexiang_2024}. However, the lesions of DR vary significantly in size and distribution, and the DR grading performance of models relying only on pure CNN architectures or pure Transformer architectures reaches a bottleneck. 
the approaches using pure CNN/ViT architecture encounter the difficulties to well tackle the DR grading task, where DR lesions significantly vary in size and scatteredly distribute, and their performances reached the bottleneck.

The hybrid framework combining CNN and Transformer architectures is a potential research line for the further performance improvement on the DR grading task. The existing CNN-and-ViT hybrid frameworks \cite{Hybrid_Sadeghzadeh_2023,HiFuse_Xiangzuo_2024} surpass the pure CNN/ViT by a large margin; however, most of them directly fuse the features extracted by corresponding stages of different backbones via averaging operations or attention modules without measuring the reliabilities/uncertainties of the features for information fusion. Such a setting degrades the interpretability of hybrid framework and leads to a demand on more rational way for feature fusion between different backbones.
% Current hybrid architectures typically connect different backbones in series or use attention modules to connect different backbone networks in parallel and use a softmax layer in the header to convert logits into predicted probabilities for each category for classification purposes. However, due to the diversity of DR lesions, the features obtained from different image inputs into multiple backbones/stages are subject to uncertainty, and existing methods have limitations in accurately estimating this uncertainty. Moreover, these methods tend to be overly confident in the fused predictive output, as softmax values essentially provide only a single-point estimate of a predictive distribution. 
% \paragraph{\bf Theory of Evidence.} {Dempster-Shafer theory, also known as evidence theory, is a framework for uncertainty-based reasoning that allows the model to combine evidences from different sources and arrive at a degree of belief \cite{A_Shafer_1976,Review_Terrence_1977}. In recent research, evidence theory has been introduced to the field of deep learning for collecting information from multi-view data to form supporting evidence for each view, and the model further aggregates the evidence from a single view to provide more robust performance \cite{Reliable_Xu_2024,Trusted_Han_2023}.}
In this regards, we propose the first evidence-theory-based multi-backbone fusion framework, which effectively integrates the features extracted by different types of backbones, leveraging their respective strengths to achieve more accurate DR grading. Specifically, Dempster-Shafer theory (\emph{a.k.a.} evidence theory) \cite{A_Shafer_1976,Review_Terrence_1977} is an approach for uncertainty-based reasoning that allows the model to combine evidences from different sources and arrive at a degree of belief \cite{Reliable_Xu_2024,Trusted_Han_2023}. Based on it, we construct evidences with features from different backbones, and accordingly forms a set of opinions regarding to the uncertainties of features for fusion and the overall uncertainty for the final decision of DR grading, \emph{i.e.,} a better interpretability for our hybrid model is achieved.
% The framework is capable of capturing not only the uncertainty at each stage of the individual backbone but also the overall uncertainty, which provides a better interpretability for the fusion and decision-making of the framework. 
The proposed hybrid framework is evaluated on two publicly available DR datasets. The experimental results demonstrate the effectiveness of our hybrid model, \emph{i.e.,} a new state-of-the-art is achieved.

\section{Method}
\begin{figure}
\includegraphics[width=\textwidth]{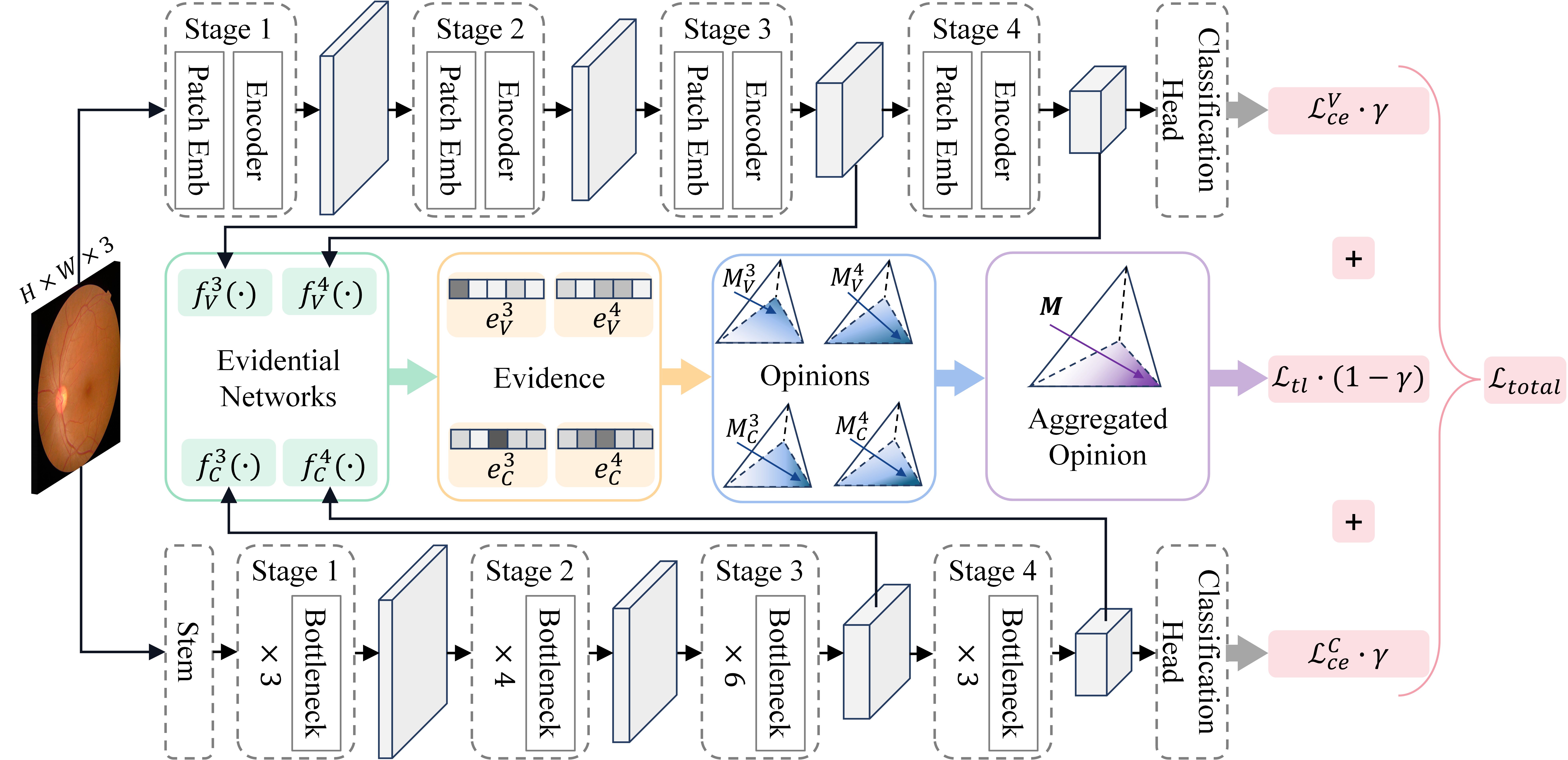}
\caption{The overall flowchart of proposed framework of multi-backbone fusion based on evidence theory. Evidences and opinions are constructed based on the features extracted by different stages of CNN and ViT, which are then adopted for feature fusion. The fusion of last two stages of CNN and ViT is taken as an example for illustration.} \label{fig:overview}
% \vspace{-5mm}
\end{figure}

The proposed hybrid framework bridging CNN and ViT based on evidence theory is illustrated in Fig.~\ref{fig:overview}.
% overall architecture leverages the strengths of different backbone networks through evidence theory, as illustrated in 
Given an input three-channel fundus image $I \in \mathbb{R}^{H \times W \times 3}$, where $H$ and $W$ denote the height and width of the image, respectively. The features extracted by different stages of CNN-based backbone and ViT-based backbone can be formulated as $F_{s}^{C}$ and $F_{s}^{V}$, respectively, where $s \in [1,\dots,4]$.
% After the CNN-based backbone \cite{Deep_He_2016} and Transformer-based backbone \cite{PVT_Wang_2022}, we can get the feature maps of the outputs of each stage of ResNet and PVT, $F_{s}^{R},F_{s }^{P},s \in [1,\dots,4]$. 
Based on the extracted features, the evidences and opinions are accordingly constructed for the reliable feature fusion, which fully integrate the strengths of different backbones. In the followings, we will introduce the construction of evidences and opinions in details.
% allow different backbone networks to fully utilize their respective strengths
% We use multiple evidence networks to collect supporting evidence from these feature maps. The distributions of the class probabilities are then modeled with Dirichlet distributions and parameterized with individual stages of evidence. From these distributions, we can construct opinions consisting of belief mass vectors and decision reliability. Finally, a reliable integrated opinion is formed by opinion aggregation. In order to allow different backbone networks to fully utilize their respective strengths, we introduce a dynamic adjusting parameter to organically combine the original backbone network training with trusted evidence learning to form an end-to-end framework.

\subsection{Construction of Evidences and Opinions}
% \vspace{1mm}
\noindent{\bf Evidences.} For the problem of $K$-classes classification (\emph{i.e.,} DR grades in this study), the $k^{th}$ class corresponds to a belief mass ($b_{k}$, $k \in [1,\dots,K]$). Then for all classes, there is $\pmb{b}=[b_{1},\dots,b_{k}]$, as well as an overall uncertainty mass $u$. The belief mass $b_{k}$ of a class $k$ is computed from the evidence of that class. Similar to \cite{Reliable_Xu_2024}, we implement respective evidential neural network $f^{n}(\cdot)$ to each stage of backbone network to collect evidence $e^{n}$, $n \in [1,\dots,N]$, where $N$ is the number of stages in the backbone. For $n^{th}$ stage, let $e_{k}^{n} \geq 0$ represent the evidence for the $k^{th}$ class, then the belief mass $b_{k}^{n}$ and uncertainty $u^{n}$ can be calculated by the following equation: 
\begin{equation}
b_{k}^{n}=\frac{e_{k}^{n}}{S^{n}},u^{n}=\frac{K}{S^{n}}, \text{and }u^{n}+\sum_{k=1}^{K}{b_{k}^{n}}=1,
\label{eq:bu}
\end{equation}
where $u^{n} \geq 0$, $b_{k}^{n} \geq 0,k \in [1,\dots,K]$, and $S^{n}=\sum_{i=1}^{K}(e_{i}^{n}+1)$. Particularly, uncertainty is inversely proportional to total evidence. When there is no evidence, each class has a belief of 0 and an uncertainty of 1. 

\vspace{1mm}
\noindent{\bf Opinions.} A belief mass distribution corresponds to a Dirichlet distribution with parameter $\alpha_{k}^{n}=e_{k}^{n}+1$. That is, the subjective opinion $b_{k}^{n}=(\alpha_{k}^{n} -1)/S^{n}$ can be easily obtained from the parameters of the corresponding Dirichlet distribution, where $S^{n}=\sum_{i=1}^{K}{\alpha_{i}^{n}}$ is termed the Dirichlet strength. The Dirichlet distribution parameterizing the evidence represents the density of each probability assignment. It models second-order probability and uncertainty.The Dirichlet distribution is a probability density function of the possible values of the probability mass function $\pmb{p}$. It is characterized by $K$ parameters $\pmb{\alpha} = [\alpha_{1},\cdots,\alpha_{K}]$: 
\begin{equation}
D(\pmb{p}|\pmb{\alpha})=\left\{
\begin{matrix}
 \frac{1}{B(\pmb{\alpha})}\prod_{i=1}^{K}{p_{i}^{\alpha_{i}-1}} & \text{for } \pmb{p} \in S_{K},\\
0 & \text{otherwise},
\end{matrix}
\right.
\end{equation}
where $S_{K}=\{\pmb{p}|\sum_{i=1}^{K}{p_{i}}=1 \text{ and } 0\leq p_{1},\cdots,p_{K}\leq 1\}$ is a $K$-dimensional unit simplex and $B(\pmb{\alpha})$ is a $K$-dimensional multinomial beta function. From eq.(\ref{eq:bu}), it can be easily deduced that the more evidence in the $k^{th}$ class, the higher the assigned belief mass. Correspondingly, the less total evidence obtained, the higher the overall uncertainty of the classification. belief assignment can be viewed as a subjective opinion. Given a subjective opinion, the mean of the corresponding Dirichlet distribution $p^{n}$ for the class probability $p_{k}^{n}$ is computed as $p_{k}^{n}=\alpha_{k}^{n}/S^{n}$.

\subsection{Multi-backbone Fusion with Trusted Evidence}
Based on the evidence theory previously mentioned, we can obtain the opinions of different stages and the corresponding class distributions. In order to fully leverage the semantic information extracted by different backbone networks and different stages for accurate DR grading, inspired by \cite{Trusted_Han_2023,Reliable_Xu_2024}, we propose to fuse the opinions based on trusted evidences. Let $M^{1}_C=(\pmb{b}^{1}$, $u^{1}$, $\pmb{a}^{1})$ and $M^{2}_C=(\pmb{b}^{2}$, $u^{2}$, $\pmb{a}^{2})$ be the opinions of stage 1 and 2 from the CNN branch, respectively, as an example. The aggregated opinions can be calculated by the following equation: 
\begin{equation}
M^{1}_C\oplus M^{2}_C=(\pmb{b}^{1 \oplus 2},u^{1 \oplus 2},\pmb{\alpha}^{1 \oplus 2})=(\frac{b^{1}_{k}u^{2}+b^{2}_{k}u^{1}}{{u^{1}+u^{2}}},\frac{2u^{1}u^{2}}{u^{1}+u^{2}},\frac{\alpha^{1}_{k}+\alpha^{2}_{k}}{2}).\label{fusion}
\end{equation}
Such a combination is achieved by mapping belief opinions to evidence opinions using a bijective mapping between the multinomial opinion and the Dirichlet distribution. The new opinion after integration satisfies that when the uncertainty of both opinions is high, the combination uncertainty will also be high, and conversely, when both opinions have low uncertainty, the final result may have high confidence. For $N$ stages, these beliefs from different stages can be combined according to Eq.~\ref{fusion} to get the final joint opinion of multiple backbones and multiple stages $\pmb{M}=M_{C}^{1}\oplus M_{C}^{2}\oplus \cdots \oplus M_{C}^{N} \oplus M_{V}^{1}\oplus M_{V}^{2}\oplus \cdots \oplus M_{V}^{N}$, which yields the combined probability and overall uncertainty of each class.

\subsection{Loss Functions}
In this section, we will introduce the loss functions adopted for the training of our evidence-theory-based hybrid framework.
% Further, we will describe training an evidential neural network to form joint opinions with multiple backbones. 
% Given a sample $x_{i}$, let $e=f(x_{i}|\theta)$ denote the evidence vector predicted by the network to be used for classification, where $\theta$ is a network parameter. 
For an input sample $x$, our hybrid model can yield the aggregated opinion $\pmb{M}$, which corresponds to an aggregated Dirichlet distribution $D(\pmb{p}|\pmb{\alpha})$, and its mean value (\emph{i.e.,} $\pmb{\alpha}/S$) can be used as an estimate of the classification probability. To calculate the loss using this estimate and ground truth, we adjust the common cross-entropy loss as: 
% \begin{equation}
\begin{align}
\mathcal{L}_{ace}(\pmb{\alpha})&=\int{{{ \left[ \sum_{j=1}^{K}{-y_{j}\log(p_{j})} \right]} \frac{1}{B(\pmb{\alpha})}\prod_{j=1}^{K}{p_{j}^{\alpha_{j}-1}d\pmb{p}}}}=\sum_{j=1}^{K}{y_{j} \left( \psi(S)-\psi(\alpha_{j}) \right)},
\end{align}
% \begin{align}
% \mathcal{L}_{ace}(\theta)&=\int{{{ \left[ \sum_{j=1}^{K}{-y_{ij}\log(p_{ij})} \right]} \frac{1}{B(\pmb{\alpha_{i}})}\prod_{j=1}^{K}{p_{ij}^{\alpha_{ij}-1}d\pmb{p_{i}}}}} \nonumber \\
% &=\sum_{j=1}^{K}{y_{ij} \left( \psi(S_{i})-\psi(\alpha_{ij}) \right)},
% \end{align}
% \end{equation}
where $y$ is a one-hot vector encoding the ground-truth class of the observation $x$, and $\psi(\cdot)$ is the digamma function. Furthermore, a Kullback-Leibler (KL) scatter term is incorporated into the loss function to guarantee that the evidence generated by the incorrect labels is lower, which forms a new loss function (\emph{i.e.,} evidence-based cross-entropy, \emph{ece}): 
% \begin{equation}
% \mathcal{L}_{acc}(\theta)=\mathcal{L}_{ace}(\theta) + \lambda_{t}{KL[D(\pmb{p_{i}}|\pmb{\tilde{\alpha}_{i}}) \Vert D(\pmb{p_{i}}|\langle 1,\cdots,1 \rangle)]},
% \end{equation}
\begin{equation}
\mathcal{L}_{ece}(\pmb{\alpha})=\mathcal{L}_{ace}(\pmb{\alpha}) + \lambda_{t}{KL[D(\pmb{p}|\pmb{\tilde{\alpha}}) \Vert D(\pmb{p}|\langle 1,\cdots,1 \rangle)]},
\label{EQ:ece}
\end{equation}
where $\lambda_{t}=\min(1.0,t/10) \in [0,1]$ is the annealing coefficient that allows the neural network to explore the parameter space, $t$ is the index of the current training epoch, $D(\pmb{p}|\langle 1,\cdots,1 \rangle)$ is the uniform Dirichlet distribution, and finally  $\pmb{\tilde{\alpha}}=\pmb{y}+(1-\pmb{y}) \odot \pmb{\alpha}$ is the Dirichlet parameter after removing non-misleading evidence from the predicted parameter $\pmb{\alpha}$ of sample $x$. In order to ensure the consistency of results between different opinions during training, minimizing the degree of confict between opinions \cite{Reliable_Xu_2024} was adopted: 
\begin{equation}
\mathcal{L}_{con}=\frac{1}{N-1}\sum^{N}_{i=1}{\sum^{N}_{i \neq j}(\frac{\sum_{k=1}^{K}|p_{k}^{i}-p_{k}^{j}|}{2}\cdot (1-u^{i})(1-u^{j}))}.
\end{equation}
In summary, the overall loss function for trusted evidence learning is as follows: 
% \begin{equation}
% \mathcal{L}_{tl}=\mathcal{L}_{acc}(\theta)+\sum_{i=1}^{N}{\mathcal{L}_{acc}(\theta^{i})}+\mathcal{L}_{con}.
% \end{equation}
\begin{equation}
\mathcal{L}_{tl}=\mathcal{L}_{ece}(\pmb{\alpha})+\sum_{i=1}^{N}{\mathcal{L}_{ece}(\pmb{\alpha}^{i})}+\mathcal{L}_{con}.
\end{equation}

% \vspace{1mm}
\noindent{\bf Overall Objective.} In our hybrid framework, the CNN and ViT branches are trained using the cross-entropy loss function, respectively. In order to fully exploit and assemble the strengths on feature extraction of different backbones,
% enable the different backbone networks to give full play to their excellent properties, 
an exponential decay strategy is adopted to integrate the training of the original backbone networks and the trusted evidence learning. The joint training objective can be written as:
% By combining the two with a parameter $\gamma \in (0,1)$ that is dynamically adjusted with the training epoch, the overall end-to-end training loss function is obtained as follows: 
\begin{equation}
\mathcal{L}_{total}= (1-\gamma) \cdot \mathcal{L}_{tl} + \gamma \cdot (\mathcal{L}_{ce}^{V}+\mathcal{L}_{ce}^{C}),
\end{equation}
where $\mathcal{L}_{ce}^{V}$ and $\mathcal{L}_{ce}^{C}$ denote the corresponding cross-entropy loss of ViT and CNN branches, respectively; $\gamma \in (0,1)$ is the loss weight tuning the relationship between branch training and evidence-based fusion learning. During the training,  $\gamma$ will gradually decay to $(1-\gamma)$ for dynamical adaptation of loss weights. Particularly, a larger $\gamma$ can ensure that each of the backbone branches well obtain the capacity of feature extraction with the supervisions of $\mathcal{L}_{ce}^{V}$ and $\mathcal{L}_{ce}^{C}$ at the early phase of training, and then focus on the feature fusion based on trusted evidences under the supervision of $\mathcal{L}_{tl}$ at the late training phase.
% Particularly, given an initial value of $\gamma$, it gradually decays to $(1-\gamma)$ with training. When the initial value of $\gamma$ is set to a larger value, it can ensure that the backbone network can carry out feature extraction and optimisation more independently at the early stage of training, and focuses on the fusion of trusted evidence with multiple backbones at the later stage of training.

\section{Experiments}
\subsection{Datasets \& Training Details}
\paragraph{\bf Datasets.} We evaluated our proposed method on the APTOS \cite{APTOS_Karthik_2019} and the DRTiD  \cite{Cross_Hou_2022}. The APTOS dataset consists of 3,662 fundus photographs collected from Aravind Eye Hospital in rural areas of India.
% is composed of fundus photographs collected by technicians from Aravind Eye Hospital in rural areas of India. 
We have divided these annotated images into training, validation, and test sets in a ratio of 7:1:2.
% , which include 2563, 366, and 733 images respectively. 
The DRTiD dataset comprises a total of 3,100 macula-centric and optic disc-centric fundus images. The dataset is divided into training set, validation set, and test set based on patients accoding to the ratio of 7:1:2, which contain 2,000, 370, and 730 images, respectively.
% We have utilized a portion of the original testing set, divided by patients, as our validation set. Following this division, the training set, validation set, and testing set now contain 2000, 370, and 730 images, respectively. 
Both datasets encompass five categories: No DR (NDR), mild DR (Mild), moderate DR (Moderate), severe DR (Severe), and proliferative DR (PDR).

\vspace{1mm}
\noindent{\bf Implementation Details.} The widely-used ResNet-50 \cite{Deep_He_2016} and PVT v2-B2 \cite{PVT_Wang_2022} are adopted as the backbones for CNN and ViT branches, respectively. Our proposed method is implemented based on the PyTorch and trained on one NVIDIA RTX A6000 GPU. We employed the stochastic gradient descent optimizer to optimize the model parameters. 
% We chose PVT v2-B2 and ResNet-50 as our backbones and initialized the parameters using models pre-trained on ImageNet. 
The initial learning rate was set to 0.001 and dynamically adjusted using a polynomial decay strategy with a maximum training epochs of 500. Consistent data augmentation settings were applied across all model training, including random cropping and random noise. 
% \paragraph{\bf Evaluation Metrics.} 
We evaluate the performance of all methods in DR grading using accuracy and quadratic weighted Kappa (Kappa).

\subsection{Comparison with State-of-the-Art}
We compared our method with three recent hybrid methods combining CNN and ViT, four popular CNN methods, and one classic ViT method. \textbf{Hybrid methods:} HiFuse \cite{HiFuse_Xiangzuo_2024} excels in various medical image classification tasks by integrating semantic information between features of different scales across multiple branches. STViT \cite{Vision_Huaibo_2023} is a hierarchical ViT hybrid with convolutional layers, demonstrating strong performance across a range of visual tasks. SMT \cite{Scale_Lin_2023}, an evolutionary hybrid network, effectively simulate the shift from capturing local to global dependencies as the network deepens, thereby achieving superior performance. \textbf{Pure CNN methods:} MPLNet \cite{MPLNet_Yining_2024} is a multi-task supervised progressive learning method that leverages DR identification task to enhance the performance of DR grading. VanillaNet \cite{VanillaNet_Chen_2023} is a carefully crafted pure CNN method, known for its simplicity and efficiency. ResNet \cite{Deep_He_2016} introduced the concept of residual learning and has been widely applied in various visual tasks. MSBP \cite{Multi_Vuong_2022}, building upon ResNet, utilizes multi-scale features in a cooperative and discriminative manner to further improve learning capabilities. \textbf{Pure ViT method:} PVT v2 \cite{PVT_Wang_2022} is an enhanced version of PVT, which not only reduces computational complexity but also improves performance on visual tasks. PVT v2 of different model sizes are involved for comparison. 

\begin{table}
\centering
\caption{Performances of different methods for diabetic retinopathy grading. The best performer is marked in \textbf{bold}, and the runner-up is marked with \underline{underline}.}
\label{tab:sota}
\begin{tabular}{l|c|c|c|c|c}
\hline
\multirow{2}{*}{\textbf{Method}} & \multirow{2}{*}{\textbf{Params (M)}} & \multicolumn{2}{c|}{\textbf{APTOS}} & \multicolumn{2}{c}{\textbf{DRTiD}} \\ \cline{3-6} 
 &  & \textbf{Accuracy} & \textbf{Kappa} & \textbf{Accuracy} & \textbf{Kappa} \\ \hline
\multicolumn{6}{l}{\bf Pure CNN} \\ \hline
MPLNet & 134.34 & 0.7981 & 0.8553 & 0.5767 & 0.5243 \\
VanillaNet-6 & 51.04 & 0.7804 & 0.8443 & 0.4795 & 0.3918 \\
MSBP & 30.02 & 0.8104 & 0.8644 & 0.6479 & 0.6679 \\
ResNet-50 & 23.52 & 0.8076 & 0.8701 & \underline{0.6603} & 0.6482 \\\hline
\multicolumn{6}{l}{\bf Pure ViT} \\ \hline
PVT v2-B3 & 44.73 & 0.8349 & 0.8887 & 0.6014 & 0.6370 \\
PVT v2-B2 & 24.85 & \underline{0.8363} & \underline{0.8957} & 0.6356 & \underline{0.6854} \\\hline
\multicolumn{6}{l}{\bf Hybrid} \\ \hline
HiFuse-Tiny & 119.69 & 0.7599 & 0.7891 & 0.4644 & 0.3016 \\
SMT-Large & 79.82 & 0.8035 & 0.8733 & 0.6151 & 0.6430 \\
STViT-Base & 50.68 & 0.8240 & 0.8826 & 0.5658 & 0.6209 \\
{\itshape Ours} & 48.39 & \textbf{0.8390} & \textbf{0.9106} & \textbf{0.6781} & \textbf{0.7118} \\ \hline
\end{tabular}
\end{table}

Table~\ref{tab:sota} shows the performances of our hybrid model and the benchmarking frameworks on the two experimental datasets. 
% Existing hybrid methods (\emph{i.e.,} HiFuse, ) are observed to yield even lower accuracy, compared to the pure CNN and ViT approaches.
Our hybrid method achieves the best performances on both datasets. In contrast, some existing hybrid frameworks (\emph{e.g.,} HiFuse) are observed to yield even lower accuracy, compared to the pure PVT v2, due to the improper feature fusion. Specifically, on the {\bf APTOS} dataset, our hybrid model improves the Kappa score by $+1.49\%$, compared to the runner-up (\emph{i.e.,} PVT v2-B2). On the {\bf DRTiD} dataset, improvements of $+1.78\%$ and $+2.64\%$ on accuracy and Kappa are yielded by our method, compared to the second-best performer ResNet-50 and PVT v2-B2, respectively. Furthermore, our approach only costs a few of extra network parameters, compared to other hybrid frameworks, which is easier for training and implementation.

\subsection{Ablation Study}
\begin{table}[]
\centering
\caption{Performances of our model with different fusion methods.}
\begin{tabular}{l|c|c|c|c}
\hline
{\multirow{2}{*}{\textbf{Method}}} & \multicolumn{2}{c|}{\textbf{APTOS}} & \multicolumn{2}{c}{\textbf{DRTiD}} \\ \cline{2-5}
& \textbf{Accuracy} & \textbf{Kappa}  & \textbf{Accuracy} & \textbf{Kappa} \\
\hline
SE & 0.8267 & 0.8787 & 0.6534 & 0.6751 \\
CBAM & 0.8226 & 0.8804 & 0.5973 & 0.6449 \\
SimAM & 0.8240 & 0.8907 & 0.6411 & 0.6586 \\
{\itshape Ours} & \textbf{0.8390} & \textbf{0.9106} & \textbf{0.6781} & \textbf{0.7118} \\
\hline
\end{tabular}
\label{tab:fusion}
\end{table}

\begin{table}[]
\centering
\caption{Performances of our model with different loss parameter settings.}%
% ED: Exponential decay strategy. CE: Cross entropy loss.
\begin{tabular}{l|c|c|c|c}
\hline
\multirow{2}{*}{\textbf{$\gamma$}} & \multicolumn{2}{c|}{\textbf{APTOS}} & \multicolumn{2}{c}{\textbf{DRTiD}}  \\ \cline{2-5}
& \textbf{Accuracy} & \textbf{Kappa}  & \textbf{Accuracy} & \textbf{Kappa} \\ \hline
0.2 & 0.8336 & 0.8994 & 0.6219 & 0.6742 \\
0.4 & 0.8240 & 0.8959 & 0.6192 & 0.6555 \\
0.6 & 0.8363 & 0.8997 & 0.6493 & 0.6859 \\
0.8 & \textbf{0.8390} & \textbf{0.9106} & \textbf{0.6781} & \textbf{0.7118} \\ \hline
% w/o ED & 0.8308 & 0.8989 & 0.6219 & 0.6800 \\
w/o CE & 0.8308 & 0.8980 & 0.6699 & 0.7087 \\
\hline
\end{tabular}
\label{tab:loss}
\end{table}

\begin{figure}[!t]
\includegraphics[width=\textwidth]{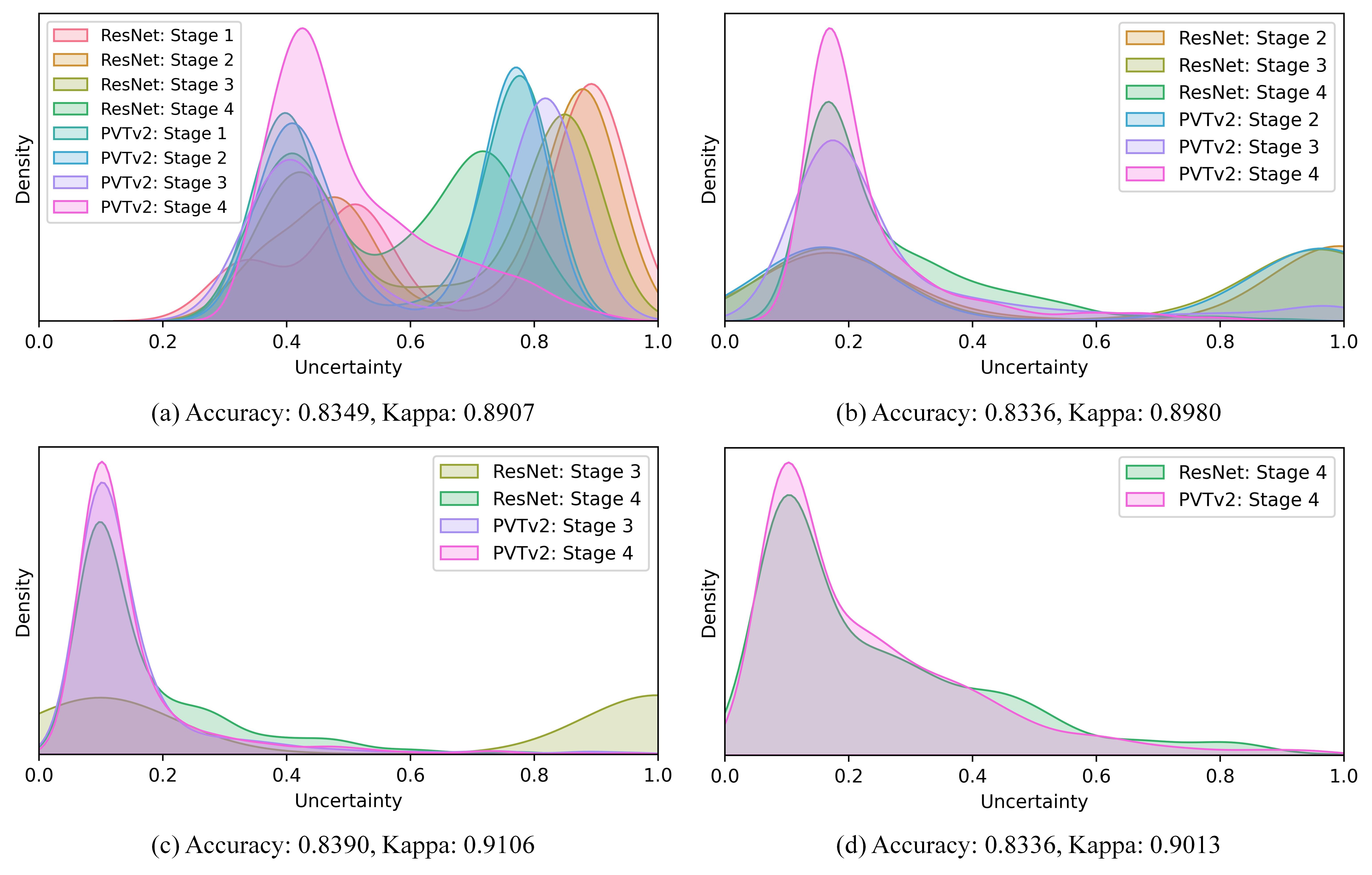}
\caption{Density of uncertainty of features yielded by different stages on APTOS.}
\label{fig:uncertainty}
% \vspace{-5mm}
\end{figure}

To demonstrate the effectiveness of evidence-theory-based fusion, three different modules are implemented to fuse features extracted by ResNet-50 and PVT v2-B2, which include squeeze-and-excitation (SE) module \cite{Squeeze_2020_Hu}, convolutional block attention module (CBAM) \cite{CBAM_Woo_2018} and simple attention module (SimAM) \cite{SimAM_Yang_2021}. The comparison results in Table~\ref{tab:fusion} show that our trusted-evidence-learning approach can effectively fuse features from multiple stages in a complementary manner, significantly surpassing the DR grading performances yielded by other fusion methods. Table~\ref{tab:loss} presents the performances of our model with different settings of loss functions. Without cross-entropy (CE) losses for branch learning  (\emph{i.e.,} $\mathcal{L}_{ce}^{V}$ and $\mathcal{L}_{ce}^{C}$), our trusted-evidence-learning paradigm can still guide the model in feature learning and fusion, which achieves the comparable accuracy and Kappa. As the parameter $\gamma$ increases, the early phase of training focuses more on the optimization of the backbone, while the later phase pays more attentions on feature fusion, \emph{i.e.,} the model with $\gamma=0.8$ achieves the satisfactory results.

% \subsection{Discussion: Uncertainty Analysis for Different Stage Fusion}
To demonstrate the interpretability of our method for feature fusion, we illustrate the uncertainty densities \cite{Trusted_Han_2023} of features yielded by different stages of each backbone on the APTOS test set in Fig.~\ref{fig:uncertainty}. The results indicate that our hybrid model consistently achieves the high performances fusing different numbers of stages. 
Since the later stages of the model capture the richer semantic features, which contribute more significantly to the final prediction, 
% reflected in the figure as 
a decrease in uncertainty is observed with the later stages (\emph{e.g.,} stage 3 and 4) of the backbones.
This is also the reason that our hybrid framework achieves the best performance only fusing the features from the last two stages of CNN and ViT, \emph{i.e.,} the features with high uncertainties may neutralize the useful information contained in other features, and finally degrade the effectiveness of feature fusion.
\begin{table}[]
\setlength{\tabcolsep}{5pt}
\centering
\caption{Performance with different methods on Chaoyang dataset. The best performer is marked in \textbf{bold}, and the runner-up is marked with \underline{underline}.}
\label{tab:chaoyang}
\begin{tabular}{l|c|c|c|c|c|c}
\hline
\multirow{2}{*}{\textbf{Method}} & \multirowcell{2}{\textbf{Params} \\ \textbf{(M)}} & \multicolumn{5}{c}{\textbf{Chaoyang}} \\ \cline{3-7} 
 &  & \textbf{Accuracy} & \textbf{Sensitivity} & \textbf{Specificity} & \textbf{Precision} & \textbf{F1-Score} \\ \hline
% \textbf{Method} & \textbf{Params (M)} & \textbf{Accuracy} & \textbf{Sensitivity} & \textbf{Specificity} & \textbf{Precision} & \textbf{F1-Score} \\ \hline
\multicolumn{7}{l}{\bf Pure CNN} \\ \hline
MPLNet & 134.33 & 0.8474 & 0.8077 & 0.9492 & 0.8022 & 0.8046 \\
VanillaNet-6 & 51.03 & 0.8200 & 0.7758 & 0.9391 & 0.7846 & 0.7760 \\
MSBP & 30.02 & 0.8361 & 0.7873 & 0.9454 & 0.7947 & 0.7888 \\
ResNet-50 & 23.52 & 0.8249 & 0.7654 & 0.9401 & 0.7749 & 0.7695 \\ \hline
\multicolumn{7}{l}{\bf Pure ViT} \\ \hline
PVT v2-B3 & 44.73 & 0.8397 & 0.8207 & 0.9468 & 0.8119 & 0.8115 \\
PVT v2-B2 & 24.85 & 0.8502 & 0.8280 & 0.9495 & 0.8211 & 0.8224 \\ \hline
\multicolumn{7}{l}{\bf Hybrid} \\ \hline
HiFuse-Tiny & 119.69 & 0.7841 & 0.7377 & 0.9292 & 0.7204 & 0.7252 \\
SMT-Large & 79.82 & 0.8284 & 0.7775 & 0.9423 & 0.7752 & 0.7752 \\
STViT-Base & 50.68 & \underline{0.8636} & \textbf{0.8396} & \textbf{0.9551} & \underline{0.8344} & \underline{0.8333} \\
{\itshape Ours} & 48.38 & \textbf{0.8657} & \underline{0.8384} & \textbf{0.9551} & \textbf{0.8347} & \textbf{0.8347} \\ \hline
\end{tabular}
\end{table}
% -------------
\begin{table}[]
\centering
\caption{Ablation study of our model with different fusion methods on Chaoyang dataset.}
\label{tab:chaoyang2}
\begin{tabular}{l|c|c|c|c|c}
\hline
\textbf{Method} & \textbf{Accuracy} & \textbf{Sensitivity} & \textbf{Specificity} & \textbf{Precision} & \textbf{F1-Score} \\ \hline
SE & 0.8544 & 0.8178 & 0.9511 & 0.8209 & 0.8178 \\
CBAM & 0.8495 & 0.8065 & 0.9494 & 0.8109 & 0.8077 \\
SimAM & 0.8509 & 0.8127 & 0.9503 & 0.8245 & 0.8132 \\
{\itshape Ours} & \textbf{0.8657} & \textbf{0.8384} & \textbf{0.9551} & \textbf{0.8347} & \textbf{0.8347} \\ \hline
\end{tabular}
\end{table}

\subsection{Generalization Evaluation}
We notice that our CNN-and-ViT hybrid framework is a general approach, which can be used for other medical image classification tasks. In this regard, we evaluate the proposed framework on publicly available Chaoyang dataset \cite{Hard_Zhu_2022} for pathological image classification. The evaluation results are shown in Table~\ref{tab:chaoyang} and Table~\ref{tab:chaoyang2}. The proposed hybrid framework achieves the best performances in terms of most metrics (\emph{i.e.,} accuracy, specificity, precision and F1-score), compared to the listing baselines, which demonstrate its excellent generalization capacity.

\section{Conclusion}
In this paper, we integrated different backbone networks rationally based on the theory of evidence to achieve accurate DR grading. Specifically, we extracted features containing different semantic information yielded by different stages of CNN and ViT, and accordingly construct evidences and opinions based on evidence neural networks for the estimate of DR grades. 
% This approach not only enhances grading performance but also provides interpretability for model fusion and decision-making. 
Extensive experiments were conducted on two public DR grading datasets. The experimental results demonstrated the effectiveness of our hybrid method. 
Furthermore, our proposed method is evaluated on histopathology image dataset and achieves the satisfactory results, which validate the scalability and potential of our hybrid framework for various medical image classification tasks.

\section*{Acknowledgment}
\begin{sloppypar}
This work was supported by Guangxi Natural Science Foundation (2024JJA170252), the Basic Ability Enhancement Program for Young and Middle-aged Teachers of Guangxi (2025KY0157), and Youth Science Foundation of Guangxi Medical University (GXMUYSF202512).
\end{sloppypar}
\bibliography{references}
\end{document}